\documentclass[conference]{IEEEtran}
\IEEEoverridecommandlockouts
\usepackage{cite}
\usepackage{amsmath,amssymb,amsfonts}
\usepackage{algorithmic}
\usepackage{graphicx}
\usepackage{textcomp}
\usepackage{xcolor}
\usepackage{booktabs}
\usepackage[authoryear]{natbib}

\usepackage[ruled,vlined,linesnumbered]{algorithm2e}
\DontPrintSemicolon
\SetKwInput{KwIn}{Input}
\SetKwInput{KwOut}{Output}
\SetKwFunction{GroupTalk}{GroupTalk}
\SetKwFunction{Execute}{Execute}
\SetKwFunction{UpdateState}{UpdateState}
\SetKwFunction{AggregateAndCalibrate}{AggregateAndCalibrate}
\usepackage{multirow}

\def\BibTeX{{\rm B\kern-.05em{\sc i\kern-.025em b}\kern-.08em
    T\kern-.1667em\lower.7ex\hbox{E}\kern-.125emX}}
\begin{document}

\title{ChartAgent: A Chart Understanding Framework with Tool Integrated Reasoning}

\author{
\IEEEauthorblockN{
Boran Wang\IEEEauthorrefmark{1}\IEEEauthorrefmark{4},
Xinming Wang\IEEEauthorrefmark{2}\IEEEauthorrefmark{3}\IEEEauthorrefmark{4},
Yi Chen\IEEEauthorrefmark{2}\IEEEauthorrefmark{3}\IEEEauthorrefmark{4},
Xiang Li\IEEEauthorrefmark{5},
Jian Xu\IEEEauthorrefmark{2}\IEEEauthorrefmark{3},
Jing Yuan\IEEEauthorrefmark{1},
Chenglin Liu\IEEEauthorrefmark{2}\IEEEauthorrefmark{3}
}

\IEEEauthorblockA{
\IEEEauthorrefmark{1}College of Artificial Intelligence, Nankai University, Tianjin 300350, China \\
}

\IEEEauthorblockA{
\IEEEauthorrefmark{2}University of Chinese Academy of Sciences, Beijing 100049, China \\
}
\IEEEauthorblockA{
\IEEEauthorrefmark{3}State Key Laboratory of Multimodal Artificial Intelligence Systems (MAIS),\\ Institution of Automation, Chinese Academy of Sciences, Beijing 100190, China \\
}
\IEEEauthorblockA{
\IEEEauthorrefmark{4}Zhongguancun Academy, Beijing 100094, China \\
}

\IEEEauthorblockA{
\IEEEauthorrefmark{5}State Key Laboratory for Novel Software Technology, Nanjing University, Nanjing 210023, China \\
}
}


\maketitle

\begin{abstract}
With their high information density and intuitive readability, charts have become the de facto medium for data analysis and communication across disciplines. Recent multimodal large language models (MLLMs) have made notable progress in automated chart understanding, yet they remain heavily dependent on explicit textual annotations and the performance degrades markedly when key numerals are absent. To address this limitation, we introduce ChartAgent, a chart understanding framework grounded in Tool-Integrated Reasoning (TIR). Inspired by human cognition, ChartAgent decomposes complex chart analysis into a sequence of observable, replayable steps. Supporting this architecture is an extensible, modular tool library comprising more than a dozen core tools—such as key-element detection, instance segmentation, and optical character recognition (OCR)—which the agent dynamically orchestrates to achieve systematic visual parsing across diverse chart types. Leveraging TIR’s transparency and verifiability, ChartAgent moves beyond the black-box paradigm by standardizing and consolidating intermediate outputs into a structured Evidence Package, providing traceable and reproducible support for final conclusions. Experiments show that ChartAgent substantially improves robustness under sparse-annotation settings, offering a practical path toward trustworthy and extensible systems for chart understanding.
\end{abstract}


\section{Introduction}
\label{sec:intro}

\begin{figure}[t]
  \centering
   \includegraphics[width=\linewidth]{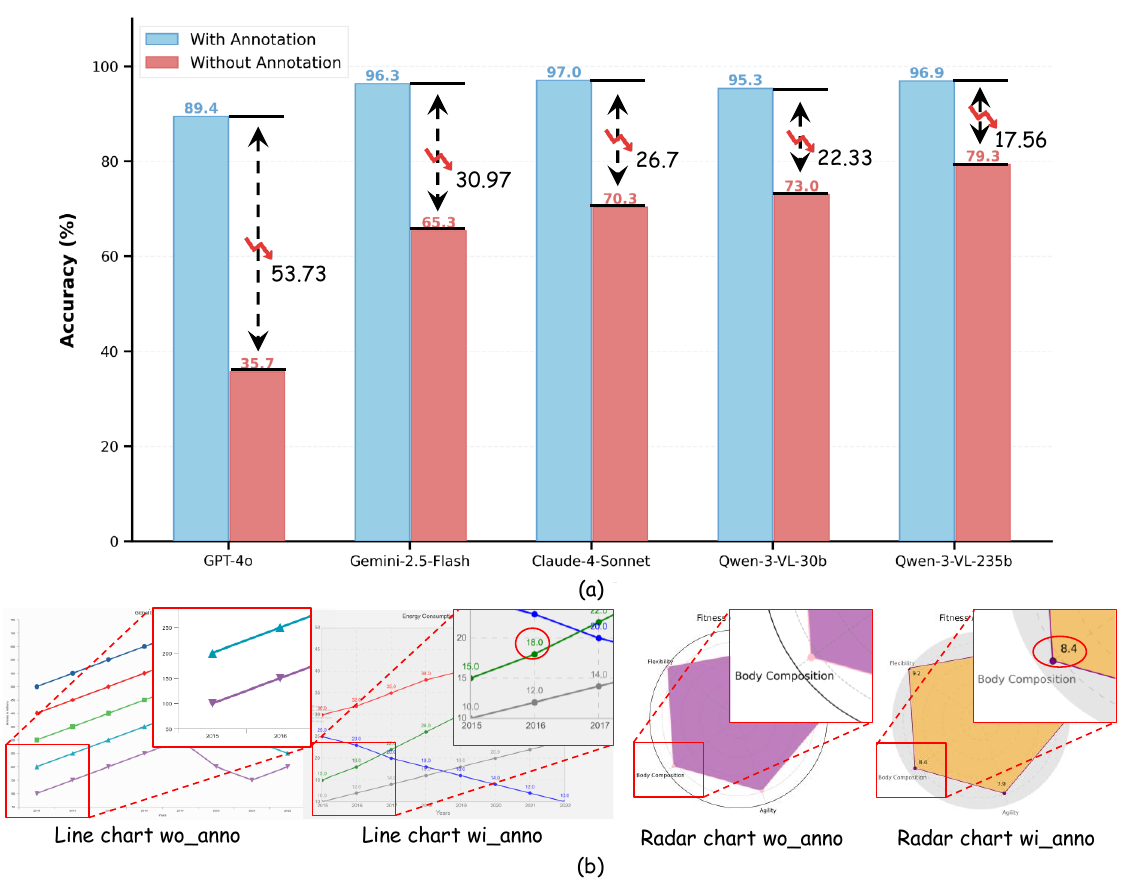}
   \caption{(a) presents the accuracy of mainstream MLLMs in answering numerical question-answering tasks on both annotated and unannotated charts. A significant decline in accuracy is observable across all models. (b) demonstrates the distinctions between the different types of annotated and unannotated charts within the ChartBench dataset. Specifically, the annotated charts feature precise numerical labels at key positions (as indicated by the content within the red circles).}
   \label{intro_fig}
\end{figure}

Charts have become indispensable to scientific research and business analytics as a universal medium for condensing data and conveying insights~\cite{wang2023scientific}. By presenting latent patterns and relationships in complex data with intuitive, high-information-density encodings, charts are among the most prevalent visualization methods in scientific literature and business reports. This widespread reliance, in turn, creates an urgent demand for automated chart understanding. In recent years, MLLMs~\cite{team2023gemini}~\cite{wang2024qwen2}~\cite{liu2023visual} have markedly advanced automated chart understanding, demonstrating strong capabilities in chart question answering, chart conversion, and data summarization. However, experiments reveal that the success of current MLLMs is highly dependent on the rich and explicit text annotations present within the chart images. Fig.\ref{intro_fig} (a) presents the performance of mainstream MLLMs on both annotated and unannotated charts from the ChartBench dataset~\cite{ChartBench}. It is evident that these models, once successful with textual cues, experience a drastic performance decline when facing real world scenarios with missing or sparse key numerical information, thereby struggling to perform accurate visual reasoning and data extraction. Specifically, GPT-4o~\cite{hurst2024gpt} exhibits an accuracy difference of 53.73\% when answering numerical questions on annotated versus unannotated charts. Even Qwen3-VL~\cite{Qwen2.5-VL}, the model least affected, observe accuracy decline by 17.56 p.p.


\begin{figure*}[t]
  \centering
   \includegraphics[width=\textwidth]{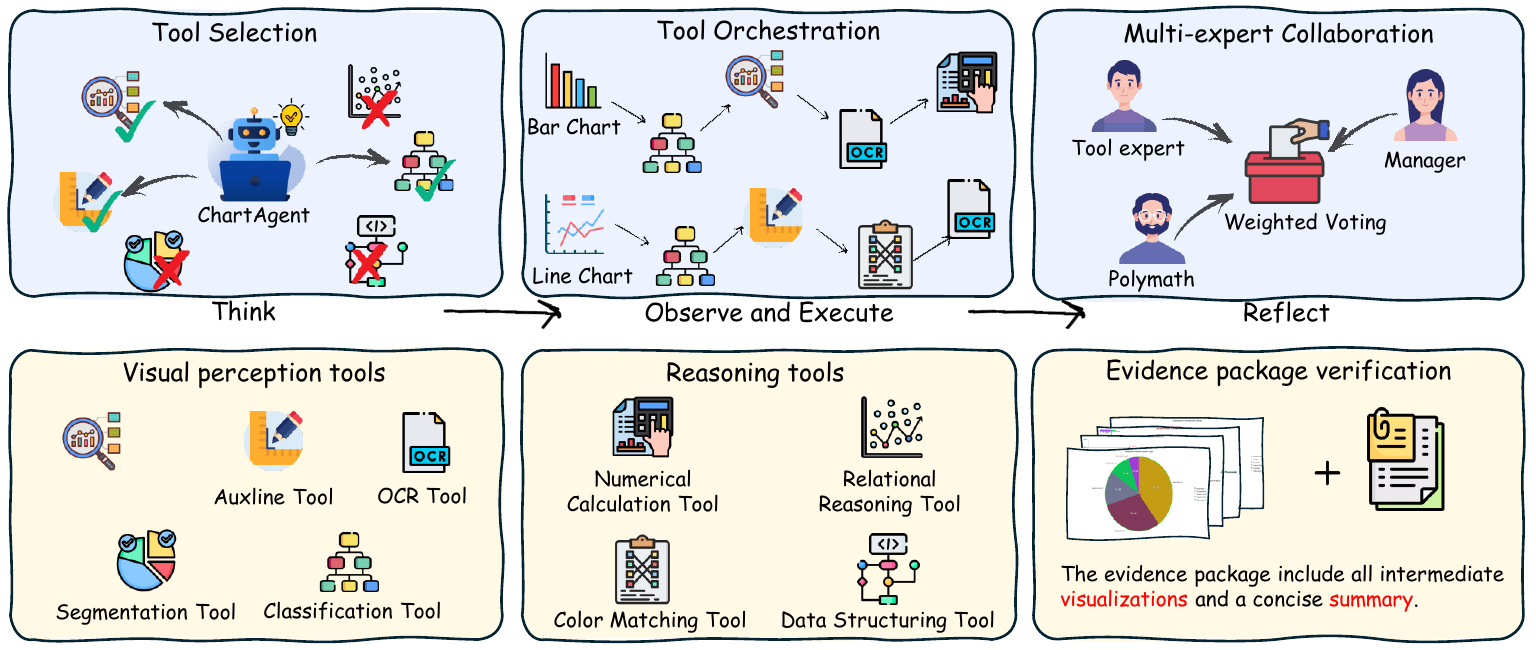}
   \caption{The figure illustrates ChartAgent's complete Think-Observe-Execute-Reflect process. The toolkit, comprising both visual perception tools and reasoning tools, support the entire process cycle of ChartAgent. The constructed evidence package subsequently offers the necessary basis for post-hoc verification.}
   \label{fig2}
\end{figure*}

This vulnerability profoundly reveals the limitations of current mainstream end-to-end models in deep visual information understanding. Most MLLMs tend to treat chart understanding as a black-box mapping from raw pixels directly to textual answers, lacking structured, verifiable intermediate visual representations and elements recognition steps~\cite{10787102}. When textual cues are insufficient, the models struggle to revert to more fundamental visual evidence (such as line chart inflection point coordinates, pie chart sector angles, etc.), thereby failing to autonomously perform reliable numerical inferences~\cite{huang2024lvlms}. More critically, this opaque reasoning paradigm undermines the traceability of the decision-making path: even when an answer is provided, researchers find it difficult to verify its basis, which consequently constrains the scalable deployment and application of chart understanding in precision-sensitive scenarios such as scientific research and finance.

In contrast, human analysts adhere to a highly structured cognitive process when interpreting charts: they do not comprehend the entire image at a single glance~\cite{salamatian2025chartgaze}, but instead adopt a decomposed and instrumentalized strategy. This involves first identifying the axes and legend, then reading the scales and labels, subsequently locating the relevant data markers based on the specific question, and completing numerical mapping through visual measurement, ultimately integrating multi-source information for calculation and reasoning. The entire procedure is step-by-step, systematic, and evidence-based.

Inspired by the human analytical process, this paper aims to overcome the core limitations of current MLLMs in comprehending charts, and proposes a chart understanding agent framework ChartAgent, based on Tool-Integrated Reasoning (TIR)~\cite{lin2025understanding}. Its core idea is to transform complex chart analysis tasks from an end-to-end black-box mapping into a reasoning chain composed of programmable steps that are observable and controllable. To support this architecture, we construct an extensible modular tool library comprising over a dozen visual and reasoning tools. ChartAgent serves as an intelligent scheduling hub, dynamically orchestrating and invoking these tools based on user queries to perform in-depth chart parsing in systematically, thereby significantly reducing reliance on textual annotations. Our main contributions are as follows:

\begin{itemize}
\item We propose a new chart-understanding agent framework, ChartAgent. The framework decomposes complex chart-reading tasks into a sequence of observable and replayable intermediate steps and dynamically orchestrates tools under a cost–gain trade-off, yielding a transparent, interpretable, and verifiable reasoning chain that provides solid evidential support for reliable conclusions.
\item A dedicated modular tool library for chart understanding is constructed. It comprises over a dozen tools, which are orchestrated and invoked by the agent. The intermediate results output by these tools are aggregated into standardized evidence packages, supporting the verifiability and robustness of the final answers.
\item ChartAgent achieves state-of-the-art(SOTA) performance with traceable interpretations across NumberQA(NQA), Value Compare, and chart-to-table tasks on the ChartQA and ChartBench datasets.
\end{itemize}

\section{Related Work}
\subsection{MLLMs for Chart Understanding}

Represented by GPT~\cite{achiam2023gpt}, Gemini~\cite{team2023gemini}, and QwenVL~\cite{Qwen-VL}, general MLLMs exhibit strong zero-shot visual reasoning owing to pre-training on massive image–text pairs. However, for chart understanding tasks that demand higher precision and stronger domain-specific priors, they remain inadequate. Their primary limitation is a text-first reasoning paradigm that over-relies on explicit textual cues in charts (e.g., legends and data labels). In addition, the patch-based visual representations~\cite{han2022survey} and aggregation mechanisms commonly employed by these models are intrinsically weak in geometric perception and in detecting subtle, continuous scale differences, making it difficult to support axis–scale mapping and fine-grained geometric measurements of length and angle. These factors collectively constrain their utility and reliability in chart understanding.

\subsection{Domain-specific Chart Models}

To address the foregoing limitations, researchers explore domain-specific chart models. Early works such as Donut~\cite{kim2022ocr} and Pix2Struct~\cite{lee2023pix2struct}, which are end-to-end, OCR-free visual document understanding models, lay the groundwork for subsequent generative methods. Recent works improve chart–text alignment and refine fine-grained reasoning pipelines to enhance MLLMs’ chart understanding~\cite{chen2024onechart}. For example, MatCha~\cite{liu2023matcha} and UniChart~\cite{masry2023unichart} augment the Pix2Struct and Donut frameworks, respectively, with chart-specific pre-training tasks—such as MatCha’s “chart de-rendering” and UniChart’s numerical reasoning—to better capture chart structure and underlying data. ChartAssistant~\cite{meng2024chartassistant} and ChartMoE~\cite{xuchartmoe} tackle the problem through task decomposition and architectural design: the former argues that general MLLMs lack structured knowledge and therefore introduces chart-to-table pre-training to strengthen the alignment between visual elements and data substantially; the latter contends that a single connector cannot accommodate the diversity of chart-understanding tasks, adopting a mixture-of-experts (MoE) framework~\cite{shazeer2017outrageously} and training specialized experts with more diverse alignment objectives (e.g., chart-to-JSON, chart-to-code) to improve generalization. Works such as MMCA~\cite{liu2024mmc} and ChartInstruct~\cite{masry2024chartinstruct} likewise employ differentiated alignment strategies to further strengthen the model’s grasp of chart content.

On the other hand, a line of work focuses on making the reasoning chain explicit and controllable. For example, TinyChart~\cite{zhang2024tinychart} adopts Program-of-Thoughts (PoT) to strengthen numerical reasoning, while ChartSketcher~\cite{huang2025chartsketcher} draws on human cognitive processes by introducing Sketch-CoT, enabling the model to annotate charts with sketch-like markings to obtain multimodal feedback and iteratively correct its visual understanding. Although such models achieve notable performance gains via domain-specific fine-tuning, they remain fundamentally end-to-end black-box systems with the decision pathways opaque and difficult to verify. Consequently, in scenarios with sparse textual annotations and a need for precise visual measurements, their robustness and trustworthiness remain constrained.

\begin{figure*}[t]
  \centering
   \includegraphics[width=\textwidth]{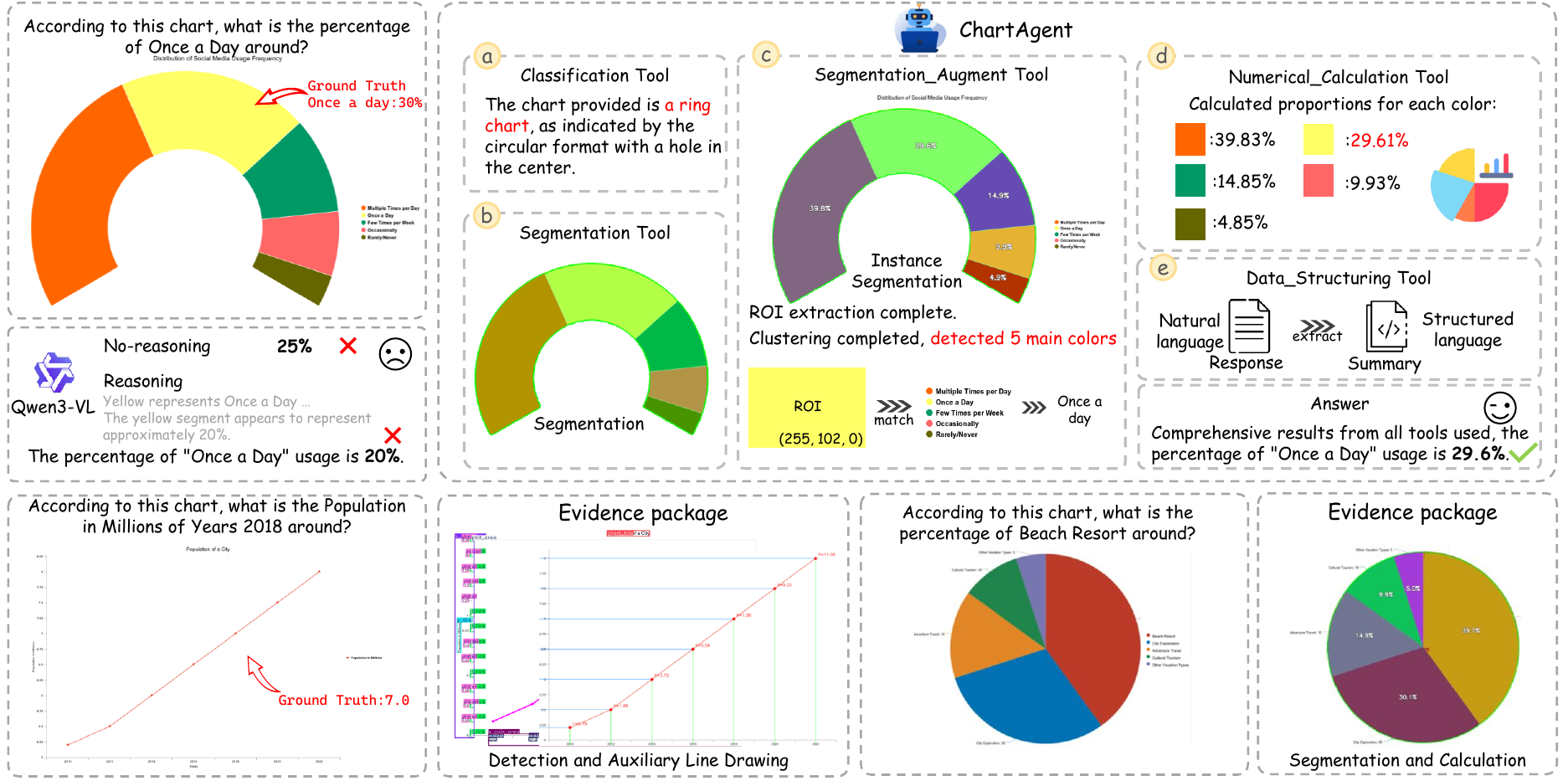}
   \caption{Two distinct chart processing scenarios are showcased. The upper panel details the tool selection and orchestration by ChartAgent when handling a sector chart lacking explicit text annotations, further displaying the intermediate results generated by each utilized tool. The lower panel showcases the resulting visualization output of the evidence package when the ChartAgent processes a line graph without annotations.}
   \label{fig3}
\end{figure*}

\subsection{Agent Frameworks with Tool Use}

In parallel with end-to-end fine-tuning, the LLM-based agent paradigm~\cite{wang2025hitchhiker}~\cite{gupta2023visual}, often referred to as tool augmentation, has gained traction~\cite{du2024anytool}. Inspired by ReAct~\cite{yao2022react}, agents decompose complex tasks into an iterative think–act–observe loop~\cite{schick2023toolformer}, so as to decouple perception from reasoning: the model delegates verifiable operations to external tools and incorporates environmental feedback to compensate for large models’ limitations in precise perception and retrieval. In chart-centric scenarios, prior approaches typically invoke specialized models such as DePlot~\cite{liu2023deplot} and StructChart~\cite{xia2023structchart} to transduce images into structured representations (e.g., tables or JSON as low-level data), then linearize these structures into text for a LLM to perform higher-level analysis and generate answers.

\section{ChartAgent}

\subsection{Overall Framework}
ChartAgent adopts a training-free, zero-shot paradigm with cognition-inspired, tool-integrated reasoning that decomposes complex chart parsing into a sequence of controllable, replayable steps. As shown in Fig. \ref{fig2}, the agent first ingests the chart image and task instruction, performs chart-type recognition and subtask decomposition, and then dynamically orchestrates the appropriate tools according to the chart type and task to progressively extract structural and semantic information. The intermediate results produced by each tool are consolidated into an evidence package, rendering the entire reasoning chain transparent, traceable, and verifiable. In the final decision stage, a GroupTalk-based multi-expert collaboration mechanism conducts consensus voting and confidence fusion over the outputs from multiple tools, producing the final answer together with a structured analysis. Leveraging the curated tool library and the large language model, the entire pipeline requires no additional training to achieve automated understanding and information extraction for complex charts.

\subsection{Modular Tool Library Construction}
To better perform chart understanding tasks, we develop an extensible, chart-oriented tool library comprising more than a dozen core capabilities, endowing the agent with two classes of functions: low-level visual perception and high-level reasoning/integration. The visual tools provide fundamental sensing—element detection, region segmentation, text recognition, and geometric measurement—while the reasoning tools undertake higher-level analysis such as numerical computation, relational modeling, and data structuring. A complete inventory of tools and additional visualization examples are provided in the Appendix.

As illustrated in the upper panel of Fig. \ref{fig3}, when a sector-type chart is given as input, ChartAgent first invokes the \texttt{ChartClassificationTool} to determine the chart type, thereby planning subsequent tool selection and orchestration. After the tools are selected, it calls the \texttt{SegmentationTool} to extract regions of interest and, together with the \texttt{SegmentationAugmentTool} tailored for pie and ring charts, performs sector-level instance segmentation. Once sufficient low-level visual evidence has been obtained, ChartAgent employs reasoning tools such as \texttt{NumericalCalculationTool} and DataStructuringTool to conduct numerical estimation, metric aggregation, and structured organization, thereby correctly answering a question previously answered incorrectly by Qwen3-VL and producing an evidence package for verification. The lower panel of Fig. \ref{fig3} showcases other representative visual tools. For line and scatter charts, the AuxLineDrawerTool leverages detected key elements to automatically draw auxiliary lines, robustly mapping data points to axis ticks and generating verifiable visual evidence. The KeyElementDetectionTool detects axes, ticks, legends, and series markers to supply reliable priors for subsequent measurement and alignment. Several tools are implemented atop mature vision models, such as YOLO~\cite{khanam2024yolov11}~\cite{yolov8_ultralytics} and SAM~\cite{kirillov2023segment}~\cite{ravi2024sam}, to establish a robust perceptual foundation.

Beyond the visual perception layer, the library also contains a set of reasoning and structuring tools to post-process and consolidate raw perceptual outputs. \texttt{NumericalCalculationTool} performs fundamental operations such as summation, normalization, ratio conversion, and logical evaluation; \texttt{RelationalReasoningTool} constructs semantic and spatial relation graphs among elements based on detection and segmentation, enabling interval comparison, trend analysis, and cross-series alignment; \texttt{DataStructuringTool} integrates fragments from OCR, detection, and computation into standardized tables or JSON so that the agent can execute final structured queries and explanations. The two classes of tools are mutually reinforcing, enabling the agent to complete a human-like “think–observe–act–reflect” chart-reading loop. All tools adhere to a unified interface schema that standardizes function descriptions, parameters, and meta-information (e.g., cost estimates and expected information gain). Building on this schema, ChartAgent performs a cost–information-gain trade-off to dynamically select the optimal tool sequence and commits outputs from each stage to a structured evidence package, thereby supporting subsequent consistency checks and auditable verification.

\subsection{Dynamic Tool Orchestration}

Before ChartAgent organizes multi-step reasoning via a function-calling mechanism, we inject the aforementioned tool interfaces and meta-information into the system prompt so that the model is aware of each tool’s functionality, parameters, and invocation cost. During iterative reasoning, the LLM emits a structured directive—either a \texttt{<tool\_call>} containing the tool name, parameters, and rationale, or a \texttt{<finish>} that terminates the reasoning. Upon receiving a \texttt{<tool\_call>}, the agent validates the tool name and parameter legality, checks invocation quotas and call limits, executes the tool, and commits its outputs to an internal evidence package. When a \texttt{<finish>} is received, the agent halts iteration, aggregates the available perceptual evidence, and proceeds to finalization. If the LLM produces unparseable or unexpected output, the system records a log and continues iterating under constraints to keep the process stable and controllable. Overall, the workflow follows a classic think–observe–act–reflect loop: the LLM undertakes decision-making and logical inference, while the tools interact with the external environment and supply verifiable observational evidence; together, they accomplish complex chart understanding.

To enable efficient and stable orchestration, the agent maintains two key caches: state and history. The state cache records concise summaries of each tool—for example, element categories and counts detected by detection tools, mask area ratios obtained from segmentation, and coordinate readings produced by auxiliary-line mappings. After compression and filtering, only salient information is retained to control prompt length and reasoning cost. The history cache logs, for every iteration, the invoked tool’s name, parameters, result summary, and call order, which supports decision tracing and error analysis while facilitating subsequent provenance tracking and verification. When constructing the next-round prompt, the system embeds the accumulated state and a pruned history so that the LLM is aware of the available information and the steps already completed, while also injecting quota reminders and cost-control policies (e.g., calls already made and recommended limits) to guide more cautious and effective decisions under a cost–information gain trade-off. Through these mechanisms, the control layer achieves fine-grained scheduling over the reasoning process, ensuring the validity and rationality of tool invocations while avoiding unnecessary repetition.

In practice, the agent typically first invokes a \texttt{ChartClassification} tool to determine the chart type, and then selects the corresponding tools to progressively acquire structured information. Once the evidence is sufficient and the answer is verifiable, the LLM outputs \texttt{<finish>} to terminate the iteration. If the reasoning process reaches the maximum number of rounds without convergence, the system forcibly enters the \texttt{<finish>} phase and triggers a fallback strategy: it returns the currently accumulated intermediate results and the evidence package, explicitly marking any incompleteness, thereby ensuring that a usable and auditable response is produced under all circumstances.

The foregoing description outlines the tool-invocation workflow from a system implementation perspective. To formalize this decision process at the computational level, we model it as a partially observable decision problem.

\paragraph{Computational formalization.}
Let the chart image and instruction be $(x,q)$ and the answer variable be $Y\!\in\!\mathcal{Y}$. At round $t$, the agent has accumulated evidence $E_t$ and maintains a belief
$b_t(y) = \Pr\!\big(Y=y \mid E_t,x,q\big)$ with entropy $H(b_t) = -\sum_{y\in\mathcal{Y}} b_t(y)\log b_t(y)$. $H(\cdot)$ denotes the Shannon entropy. The action space is $\mathcal{A}=\mathcal{T}\cup\{\textsf{finish}\}$, where $\mathcal{T}$ is the set of callable tools with nonnegative invocation cost $c(a)\!\ge\!0$ for $a\in\mathcal{T}$. Invoking a tool $a$ produces a stochastic observation $Z_a$ drawn from
\begin{align}
Z_a \sim p(\,\cdot \mid a,E_t,x,q), \\
b_{t+1}^{(a,z)}(y) = \Pr\!\big(Y=y \mid E_t,a,z,x,q\big).
\end{align}

\paragraph{Myopic decision rule under a cost-gain trade-off.}
We score each tool by its expected information gain (EIG), written compactly as a conditional mutual information:
\begin{align}
\mathrm{EIG}_t(a) 
&= I\!\big(Y; Z_a \,\big|\, E_t,x,q\big) \notag\\
&= H(b_t) - \mathbb{E}_{z\sim p(\cdot \mid a,E_t,x,q)} \!\left[\, H\!\left(b_{t+1}^{(a,z)}\right) \right].
\label{eq:eig}
\end{align}
A one-step lookahead policy chooses
\begin{align}
a_t \;=\; \arg\max_{a\in \mathcal{T}} \Big\{ \mathrm{EIG}_t(a) - \lambda\, c(a) \Big\},
\label{eq:policy}
\end{align}
where $\lambda\!>\!0$ balances information gain against cost. A budget $B$ enforces $\sum_{s=1}^{t}c(a_s)\le B$. We terminate when
\begin{align}
\max_{a\in\mathcal{T}} \big\{ \mathrm{EIG}_t(a) - \lambda c(a) \big\} \le \eta
\quad \text{or} \quad 
\sum_{s=1}^{t} c(a_s)\, \ge B,
\label{eq:stop}
\end{align}
with threshold $\eta\!\ge\!0$, and then produce the final decision based on $b_t$ together with the evidence package.

\begin{table*}[ht]
\centering
\caption{Performance of ChartAgent and prior methods on ChartBench. Relaxed accuracy (\%) for the NumberQA (NQA) task and accuracy (\%) for the Value Compare (VC) task are reported for chart-specific models and general-purpose MLLMs.}
\begin{tabular}{l c | l c | l c}
\toprule
\multicolumn{6}{c}{\textbf{ChartBench}}\\
\midrule
\multicolumn{4}{c|}{\textbf{NumberQA (NQA)}} &
\multicolumn{2}{c}{\textbf{Value Compare (VC)}}\\
\cmidrule(r){1-4}\cmidrule(l){5-6}
\textbf{Chart Models} & \textbf{Relax Acc.} &
\textbf{General MLLMs} & \textbf{Relax Acc.} &
\textbf{General MLLMs} & \textbf{Acc.} \\
\midrule
MatCha        & 25.86 & GPT-4o             & 45.33 & GPT-4o             & 65.42 \\
UniChart      & 27.44 & Gemini-2.5 Flash   & 63.26 & Gemini-2.5 flash   & 84    \\
ChartMoE      & 51.67 & Claude4            & 63.86 & Claude4            & 78.32 \\
ChartVLM     & 32.19 & Qwen2.5VL-72b      & 63.11 & Qwen2.5VL-72b      & 79.61 \\
ChartLlama    & 21.31 & Qwen3VL-30b        & 63.81 & Qwen3VL-30b        & 66.57 \\
TinyChart     & 47.86 & \textbf{CharAgent(Qwen3vl-30b)} &
\underline{68.46($\uparrow$4.65)} &
\textbf{ChartAgent} & \underline{69.78($\uparrow$13.21)} \\
ChartGemma    & 38.46 & Qwen3VL-235b       & 71.48 & Qwen3VL-235b       & 82    \\
ChartInstruct & 31.75 & \textbf{ChartAgent(Qwen3vl-235b)} &
\textbf{73.25($\uparrow$1.77)} &
\textbf{ChartAgent} & \textbf{84.57($\uparrow$12.57)} \\
\bottomrule
\end{tabular}
\label{table1}
\end{table*}

\subsection{Multi-Expert Collaboration}
ChartAgent introduces a multi-expert collaboration mechanism at the decision stage to simulate multi-party deliberation and enhance reliability and robustness under uncertainty. Concretely, we preset several experts with complementary foci who share the same historical state but emphasize different evidence dimensions: a tool expert evaluates the rationality of each tool invocation and the quality of its outputs, while a synthesis expert concentrates on integrating cross-source cues and performing consistency checks. When arbitration is required (e.g., conflicting results or dispersed confidences), each expert independently issues a vote with a score and a confidence estimate based on the current image and accumulated state. The votes are then aggregated and calibrated—performing score normalization and confidence re-scaling—and, when necessary, thresholds and temperature parameters are applied to select the top-scoring candidate as the final decision, with optional remedial invocations or additional verification triggered by rule to mitigate the risk of error.

As a pluggable decision-enhancement module, GroupTalk can be inserted at key nodes of the main reasoning loop, including tool selection (e.g., prioritizing between key-element detection and direct OCR), conflict resolution (arbitrating or triggering re-validation when tools disagree), and result fusion (weighted integration across sources or selection of the highest-confidence conclusion). This mechanism injects “collective intelligence” into ChartAgent, substantially improving robustness and transparency under complex and noisy conditions. The rationales, confidence scores, and cited evidence for all ballots—together with the aggregation and arbitration procedures—are logged step-by-step into the evidence package to form an auditable decision trail, facilitating post-hoc traceability and human verification.

\subsection{Evidence Package Verification}
To enable verifiable reasoning, ChartAgent introduces a chart evidence package (CEP) mechanism at the output stage. The CEP systematically aggregates key intermediate artifacts generated throughout the pipeline—including detected bounding boxes with visualization overlays, OCR-extracted text with coordinates, segmentation masks, color statistics and comparative results, auxiliary-line alignment diagrams, and GroupTalk voting and arbitration logs and archives them in a structured format with indices and concise summaries, providing traceable support for the final answer. With the CEP, ChartAgent is no longer an inscrutable black box: users and researchers can replay the reasoning trajectory step by step to verify that each operation is evidence-based, thereby substantially improving interpretability and trustworthiness. For example, when answering “What is the value for a specific year in the line chart?”, the CEP presents the localization of the year label, screenshots and readings from auxiliary-line–to–axis-tick mappings, and the corresponding text-extraction results, making the evidential basis of the answer explicit. The CEP likewise facilitates error diagnosis: if the final conclusion deviates, developers can quickly pinpoint the faulty tool or stage, providing direct cues for subsequent fixes and system improvements. Overall, this mechanism renders the reasoning process explicit and replayable, enabling traceable, evidence-grounded question answering and laying the groundwork for deployment in high-reliability settings.

\begin{table*}[t]
\centering
\caption{Results on ChartQA. Accuracy (\%) on augmented (Aug.) and human-authored (Human) questions, as well as their average (Avg.), is reported for chart-specific models, general MLLMs, and our ChartAgent.}
\begin{tabular}{lccc|lccc}
\toprule
\multicolumn{8}{c}{\textbf{ChartQA}}\\
\midrule
\textbf{Chart Models} & \textbf{Aug.} & \textbf{Human} & \textbf{Avg.} &
\textbf{General MLLMs} & \textbf{Aug.} & \textbf{Human} & \textbf{Avg.} \\
\midrule
MatCha        & 86.64 & 37.12 & 61.88 & GPT-4o                      & 85.52   & 69.52   & 77.52   \\
UniChart      & 83.28 & 34.64 & 58.96 & Gemini-2.5                  & 86.72   & 78.23   & 82.45    \\
ChartMoE      & 90.96 & 78.32 & 84.64 & Claude4                     & 82.72   & 76.16   & 79.44    \\
ChartVLM      & 82.48 & 42.08 & 62.28 & Qwen2.5VL-72b               & 94.35   & 62.95   & 78.65    \\
ChartLlama    & 93.12 & 58.40 & 75.76 & Qwen3VL-30b                 & 93.20   & 66.18   & 79.69    \\
TinyChart     & 94.48 & 58.72 & 76.60 & \textbf{ChartAgent(Qwen3VL-30b)}     & 94.62   & 75.23   & 84.92    \\
ChartGemma    & 90.80 & 80.16 & 85.48 & Qwen3VL-235b                & 94.80   & 77.36   & 86.08    \\
ChartInstruct & 85.04 & 66.64 & 75.84 & \textbf{ChartAgent(Qwen3VL-235b)}    & \textbf{95.91} & \textbf{83.30} & \textbf{89.61}     \\
\bottomrule
\end{tabular}
\label{table2}
\end{table*}

\section{Experiment}
\subsection{Dataset and Evaluation Metrics}
\paragraph{Dataset} We systematically evaluate ChartAgent on two public chart-understanding datasets, ChartBench~\cite{ChartBench} and ChartQA~\cite{masry2022chartqa}. ChartBench targets charts with sparse textual annotations and spans 9 major categories and 42 subcategories; on this dataset, we focus on two relatively challenging tasks: NumberQA(NQA) and Value Compare. The ChartQA test set contains 1,250 question–answer pairs, with questions divided into Human and Augmented subsets. It covers the three most common chart types, and all charts originate from real-world scenarios. On ChartQA, we evaluate both chartQA and the chart-to-table conversion task.

\paragraph{Evaluation Metrics and Implementation Details} Similar to~\cite{wei2025chartmind}~\cite{yang2024chartmimic}, we adopt relaxed accuracy for NQA to better reflect practical tolerance in chart reading, and standard accuracy for Value Compare. For the chart-to-table task, we follow ~\cite{liu2023deplot} and report the RMS$_{F1}$ metric to ensure comparability with prior work. In this section, all ChartAgent variants adopt Qwen3VL as the base model. ChartAgent(Qwen3VL-30B) and ChartAgent(Qwen3VL-235B) denote the model-driven agent variants instantiated with the 30B and 235B base models, respectively.

\subsection{Performance Comparison}
\label{4.2}

\paragraph{ChartBench}
As shown in Table~\ref{table1}, on the ChartBench dataset ChartAgent demonstrates outstanding performance. For NQA, the lenient accuracy of the 30B and 235B variants reaches 68.46\% and 73.25\%, respectively—surpassing all compared domain-specialized chart models (ChartMoE, 51.67\%; TinyChart, 47.86\%) and exceeding their corresponding baselines Qwen3-VL-30B (63.81\%) and Qwen3-VL-235B (71.48\%) by 4.65 and 1.77 percentage points. For VC, ChartAgent likewise excels: the 30B and 235B versions achieve 69.78\% and 84.57\%, substantially outperforming other chart models and improving over the baselines (66.57\% / 82.00\%) by 3.21 and 2.57 percentage points. These results indicate that when questions rely more heavily on cross-element alignment and the integration of verifiable visual evidence, tool-integrated reasoning injects low-level perception and measurement capabilities that markedly enhance robustness.

\begin{table}[htbp]
  \centering
  \caption{Comparison of different models on Chart-to-Table.}
  \begin{tabular}{lc|lc}
    \toprule
    \textbf{Chart Models} & $\mathrm{RMS}_{F_1}$ & \textbf{General MLLMs} & $\mathrm{RMS}_{F_1}$ \\
    \midrule
    Pix2Struct & 78.9  & GPT-4o     & 82.62 \\
    UniChart   & 80.2  & Claude-4   & 90.57 \\
    Chartllama & 85.8  & Gemini2.5  & 77.98 \\
    ChartAss   & 85.62 & Qwen2.5VL  & 79.82 \\
    MatCha     & 72.45 & GPT-4v     & 81    \\
    Deplot     & 79.64 & \textbf{ChartAgent} & \textbf{91.53} \\
    \bottomrule
  \end{tabular}
  \label{table3}
\end{table}

\begin{figure}[t]
  \centering
   \includegraphics[width=0.8\linewidth]{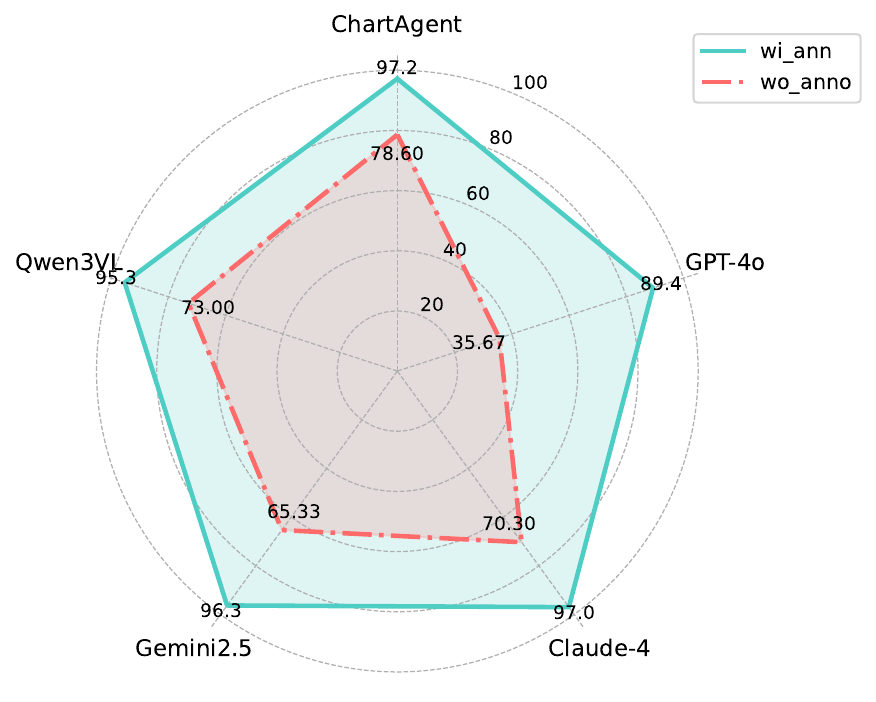}
   \caption{Accuracy (\%) of MLLMs and ChartAgent on charts with explicit numerical annotations (With ann) and on de-annotated charts without such labels (Without ann).}
   \label{radar}
   \label{fig4}
\end{figure}
\paragraph{ChartQA}
Table~\ref{table2} and Table~\ref{table3} present ChartAgent’s performance on various tasks in the ChartQA dataset. For chart question answering (QA), under the large-parameter configuration, ChartAgent attains 95.91\% / 83.30\% / 89.61\% on Aug./Human/Avg., respectively, comprehensively surpassing the baseline; notably, it achieves a substantial +5.94 p.p. gain on the Human subset, which relies more heavily on the alignment and integration of visual evidence. The smaller model, ChartAgent (Qwen3-VL-30B), likewise increases the baseline’s average accuracy from 79.69\% to 84.92\%. Consistent with the findings in Table~\ref{table1}, ChartAgent not only outperforms its size-matched MLLM baselines but also exceeds all other chart-specialized models. For the chart-to-table conversion task, ChartAgent similarly outperforms current MLLMs and several specialized systems such as ChartLlama~\cite{han2023chartllama} and ChartAssistant~\cite{meng2024chartassistant}, corroborating its strong capabilities in structured information extraction and integration.

\paragraph{Robustness to Annotation Removal}
To assess the dependence on textual annotations discussed above, we selected 300 chart–question pairs from ChartBench that contain detailed textual labels and conducted experiments with their corresponding de-annotated chart versions. As shown in Fig.~\ref{fig4}, in contrast to MLLMs whose accuracy drops sharply after annotation removal, ChartAgent is affected the least. This clearly indicates that ChartAgent substantially reduces its reliance on explicit textual annotations and corroborates its robustness in sparsely annotated settings. This robustness arises from its integrated tool library and verifiable reasoning chain. More detailed visualization results are provided in the Appendix.

\subsection{Ablation Study}

\begin{table}[t]
  \centering
  \caption{Ablation on Tool library, GroupTalk, and EIG strategy.}
  \label{tab:ablation-en-tabularstar}
  \setlength{\tabcolsep}{4pt}
  \begin{tabular}{@{\extracolsep{\fill}}lcc@{}}
    \toprule
    Variant & Acc (\%) & Tool Calls  \\
    \midrule
    w/o EIG            & 77.9 & 8.4  \\
    w/o GroupTalk      & 76.8 & 5.6  \\
    w/o Tool Library   & 75.4 & - \\
    \textbf{ChartAgent } & \textbf{78.6} & \textbf{6.1}  \\
    \bottomrule
  \end{tabular}
  \label{table4}
\end{table}

\begin{table}[t]
  \centering
  \caption{Combined EIG sweeps on ChartBench: top—cost weight $\lambda$ ($B{=}8$ fixed); bottom—budget $B$ ($\lambda{=}0.2$ fixed).}
  \label{tab:eig-combined}
  \setlength{\tabcolsep}{4pt}        
  \renewcommand{\arraystretch}{1.0}
  \begin{tabular}{c|ccc}
    \toprule
    Param & Value & Acc (\%) & Tool Calls \\
    \midrule

    \multirow{4}{*}{$\lambda$} & 0.0                 & 77.3 & 9.8 \\
                               & 0.2 (default) & 78.6 & 6.1 \\
                               & 0.5                 & 78.0 & 5.9 \\
                               & 1.0                 & 76.4 & 4.3 \\
    \cline{2-4}
    \hline
    \multirow{4}{*}{$B$}       & 3                   & 73.2 & 3.0 \\
                               & 5                   & 77.9 & 4.7 \\
                               & 8 (default)   & 78.6 & 6.1 \\
                               & 12                  & 78.8 & 7.5 \\
    \bottomrule
  \end{tabular}
  \label{table5}
\end{table}

\paragraph{Component Ablations}
We conduct ablation studies on the ChartBench subset referenced in \ref{4.2} to quantify the contribution of each ChartAgent component (Table~\ref{table4}). The full system attains 78.6\% accuracy with an average of 6.1 tool invocations. Removing the EIG-based scheduling strategy reduces accuracy to 77.9\% while increasing the average number of tool invocations to 8.4, confirming that an information-gain–driven decision rule effectively suppresses redundant calls and can even improve accuracy under a fixed budget. Eliminating the modular tool library leads to the largest degradation, with accuracy dropping from 78.6\% to 75.4\%, highlighting its foundational role as the core capability of the system. Removing the GroupTalk module also yields a notable decline (to 76.8\%), with an average of 5.6 invocations, close to the full configuration, indicating that cross-expert consensus building and confidence fusion are crucial for harder chart-reasoning cases.

\paragraph{EIG }
We further analyze the behavior of the EIG scheduler (Table~\ref{table5}). With the budget fixed at $B=8$, increasing the cost weight $\lambda$ from 0.0 to 0.2 raises accuracy from 77.3\% to 78.6\% and reduces the average number of tool invocations from 9.8 to 6.1. A large $\lambda=0.5$ further lowers invocations to 5.9 at a slight accuracy cost (78.0\%). When $\lambda$ becomes too large ($\lambda=1.0$) the scheduler turns overly conservative, producing only 4.3 invocations on average and causing accuracy to drop to 76.4\%.

Fixing $\lambda=0.2$ and varying the budget $B$ reveals a similar trade-off. An extremely small budget $(B=3)$ severely constrains the agent, yielding 73.2\% accuracy with 3.0 tool invocations. Increasing the budget to $B=5$ and $B=8$ gradually improves performance to 77.9\% and 78.6\%, with 4.7 and 6.1 invocations, respectively. Further expanding the budget to $B=12$ yields only a marginal accuracy gain (78.8\%) while incurring higher cost (7.5 invocations). Overall, we adopt $\lambda=0.2$ and $B=8$ as the default configuration, which strikes a favorable balance between answer quality and tool-invocation efficiency.

\section{Conclusions}

This paper proposes ChartAgent, which reframes chart understanding from an end-to-end black-box mapping into an observable, replayable, and verifiable reasoning chain via TIR, thereby fundamentally reducing MLLMs’ dependence on explicit textual annotations. Leveraging a modular tool library and CEPs, the system remains robust even when key numerical cues are sparse or missing, while maintaining interpretability and trustworthiness. Experimental results show that ChartAgent improves structured understanding and provides a methodological foundation for building trustworthy chart understanding system.

Future work will expand the tool library to cover richer chart elements and geometric relations, and broader application scenarios. For application, we will integrate ChartAgent into a larger-scale scientific agent ecosystem to promote the automated flow from multimodal data to verifiable knowledge, advancing chart understanding from \textit{can do} to \textit{trustworthy}.

\bibliographystyle{plainnat}

\bibliography{main}

\clearpage
\setcounter{page}{1}
\appendix


\section{Detailed of Tool Library}
\label{tool library}

\begin{table*}[t]
\centering
\caption{Overview of Tool Library} 
\label{tool}
\renewcommand{\arraystretch}{1.35}
\begin{tabular}{p{3.2cm} | p{4cm} | p{8cm}}
\toprule
\textbf{Type} & \textbf{Tools} & \textbf{Brief Explanation} \\
\midrule

\multirow{7}{*}[-136pt]{Visual perception tools} 
& Segmentation Tool &
Used to precisely \textbf{isolate specific regions or objects} (like legend items, bar segments, or pie slices) from the chart background, providing clean input for subsequent recognition and measurement.
\\ \cline{2-3}

& AugmentSegmentation Tool &Performs \textbf{instance segmentation} on the chart image to precisely isolate and identify every individual element. Following segmentation, it calculates the \textbf{pixel ratio} or area percentage occupied by each distinct instance relative to the data area or the entire chart, allowing for proportional analysis.
\\ \cline{2-3}

& AuxLine Tool &
Automatically \textbf{generates and projects horizontal or vertical auxlines} onto the chart, used to pinpoint the axis values corresponding to data points for accurate numerical reading or comparison.
\\ \cline{2-3}

& KeyElement Detection &
Specifically \textbf{identifies and localizes the core components} of the chart, such as the title, axis labels, data area, legends, and tables, serving as the starting point for structural chart understanding.
\\ \cline{2-3}

& Curve Detection &
Focuses on \textbf{recognizing and tracking data curves} or trend lines in line charts, scatter plots, etc., to extract discrete data points or analyze the change trajectory.
\\ \cline{2-3}

& OCR Tool &
Processes all textual information in the chart image, including axis tick values, titles, legend text, and data labels, converting image-based text into processable strings.
\\ \cline{2-3}

& Color Matching Tool &
Responsible for \textbf{associating the visual elements in the data area} (bars or curves) with the colors and labels defined in the legend, thereby determining each data series' meaning.
\\
\midrule

\multirow{4}{*}[-30pt]{Reasoning tools}
& Classification Tool &
Used to \textbf{identify} the overall \textbf{chart type}.
\\ \cline{2-3}

& Numerical Calculation Tool &
Executes fundamental \textbf{mathematical operations and logical inferences}.
\\ \cline{2-3}

& Data Structuring Tool &
Organizes OCR, detection, and numerical results \textbf{into a standard table or JSON format}, enabling structured queries and analysis.
\\ \cline{2-3}

& Relational Reasoning Tool &
Analyzes semantic and spatial \textbf{relationships between chart elements}, constructing a logical graph from segmentation and detection results.
\\
\bottomrule
\end{tabular}
\end{table*}

To overcome the limitations of current multimodal large language models (MLLMs) in fine-grained visual measurement and traceable reasoning, we develop a modular tool library covering more than a dozen core capabilities. Detailed tool names and descriptions are provided in Table~\ref{tool}

\begin{figure}[h]
    \centering
    \includegraphics[width=\linewidth]{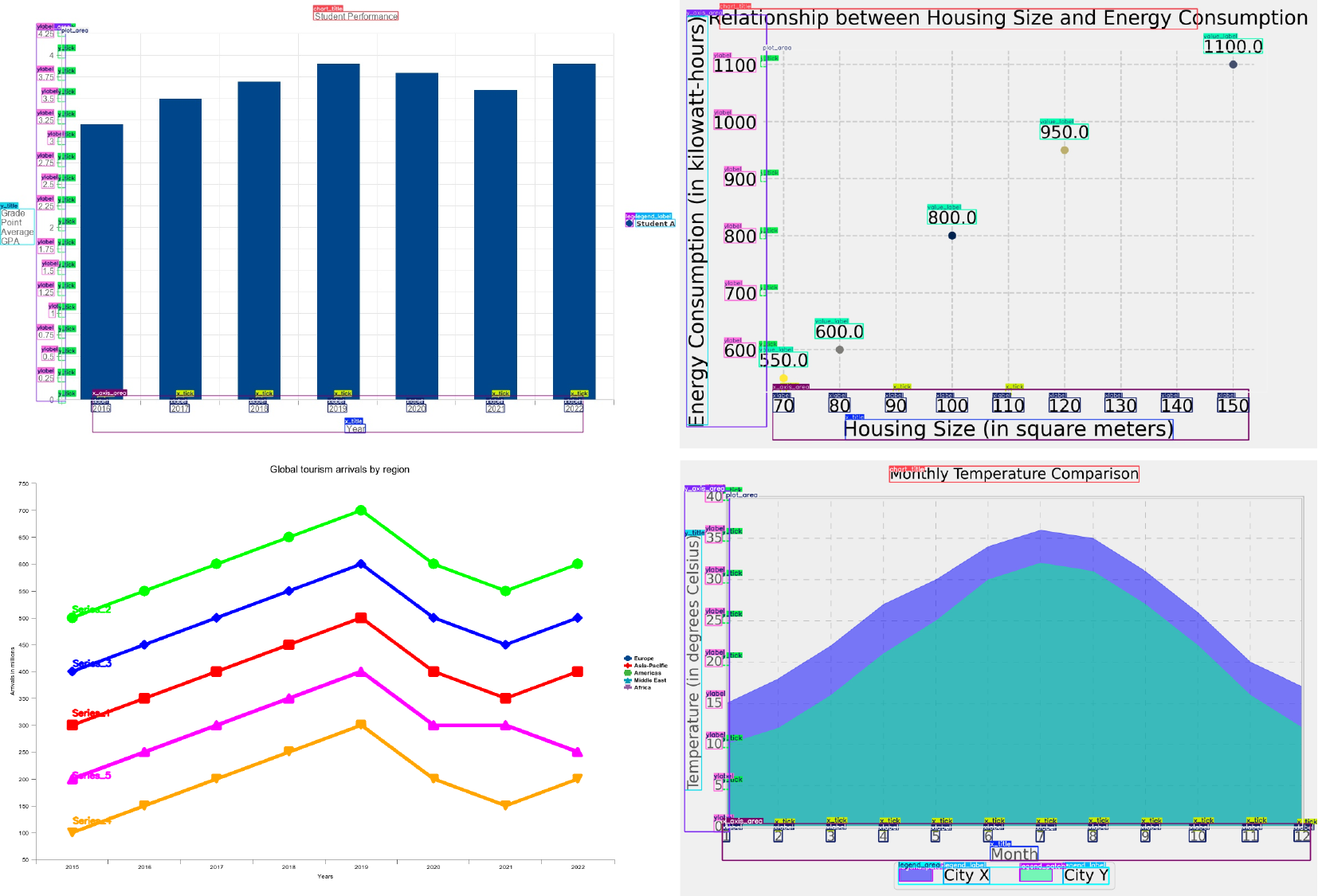}
    \caption{Low-level perception visualization results.}
    \label{app1}
\end{figure}

By coupling low-level visual perception with high-level logical reasoning, this library endows the agent with chart-analysis abilities comparable to those of a human analyst. At the perception layer, we train a dedicated YOLO-based \texttt{KeyElement Tool} to precisely localize the plot area, axes and ticks, legends, and data elements (e.g., bars) from unstructured bitmap images; its outputs not only provide spatial priors for subsequent geometric computations but also bridge the pixel and semantic spaces. The \texttt{Curve Detection Tool}, operating in the HSV color space with multi-channel thresholding and skeletonization, extracts clean topological skeletons that capture trend information from complex backgrounds. Meanwhile, textual content serves as a crucial semantic anchor for chart understanding. To this end, we integrate a Qwen-VL-OCR engine as \texttt{OCR Tool} that leverages the multimodal representation capacity of MLLM to jointly extract text content and its precise bounding boxes with confidence scores, thereby improving robustness to rotated text, dense numeric labels, and special symbols, and enabling reliable alignment between numeric values and textual annotations. Several visualization results of these tools are shown in the fig~\ref{app1}.

To accommodate different data-encoding paradigms, the tool library further incorporates specialized high-level measurement modules. For pie and donut charts, where values are expressed through area proportions, we introduce the \texttt{AugmentSegmentation Tool} following a segment–cluster–quantify pipeline: SAM’s zero-shot segmentation is first employed, using chart-type–aware prompts (e.g., a single center point for solid pies, a combination of center-negative points and bounding-box prompts for donut charts) to extract the main region. Next, valid pixels are sampled in the HSV/RGB color spaces, and K-Means clustering is applied to automatically group dominant color clusters while filtering anti-aliasing noise. Finally, the pixel proportion of each cluster is computed to infer the underlying data distribution.

\begin{figure}[h]
    \centering
    \includegraphics[width=\linewidth]{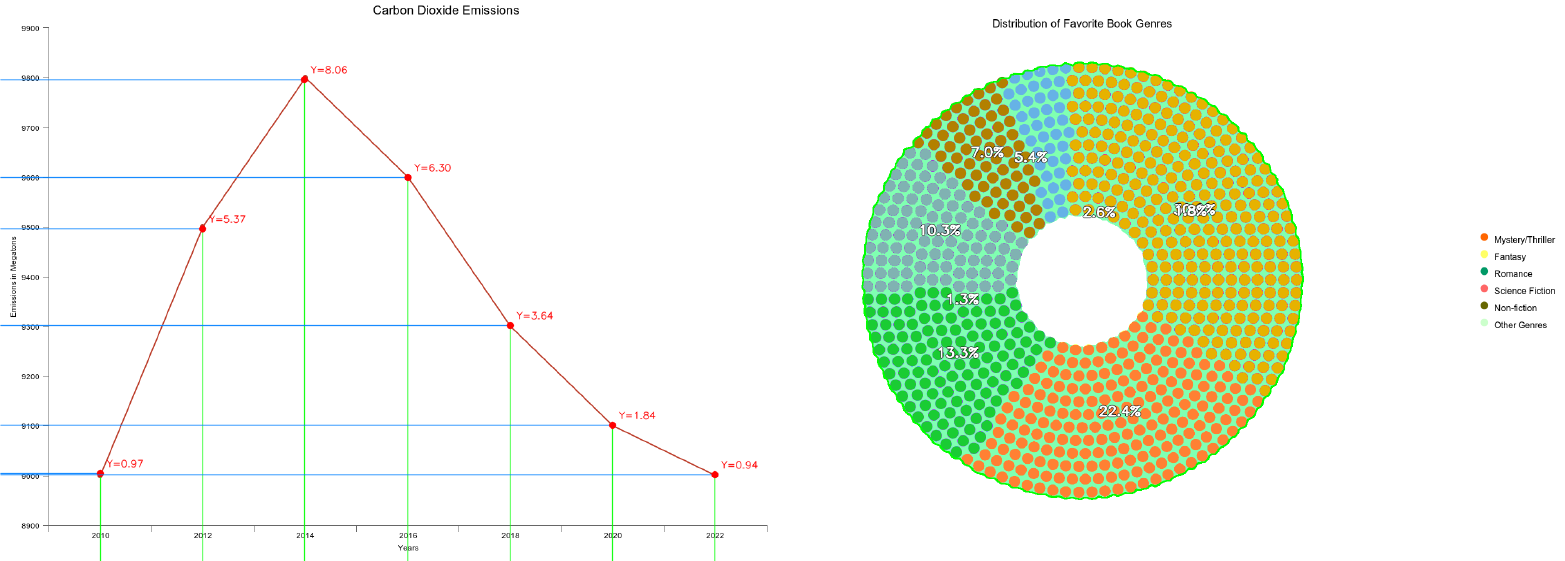}
    \caption{High-level perception visualization results.}
    \label{app2}
\end{figure}

For Cartesian-coordinate charts such as line and bar charts, we design an \texttt{AuxLine Tool} that mimics the human auxiliary-line projection strategy. Leveraging intermediate outputs from detection modules, the tool builds a calibration function between pixel coordinates and numeric values via linear regression. Using x-axis ticks as anchors, auxiliary lines are projected toward the y-axis to locate geometric intersections between projection paths and data skeletons; the calibration model then maps intersection pixel coordinates back into precise numerical values, enabling a reliable conversion from image-level measurements to structured data. Visualization results for the above tools are shown in the fig.~\ref{app2}.

\begin{figure*}[h]
    \centering
    \includegraphics[width=0.8\textwidth]{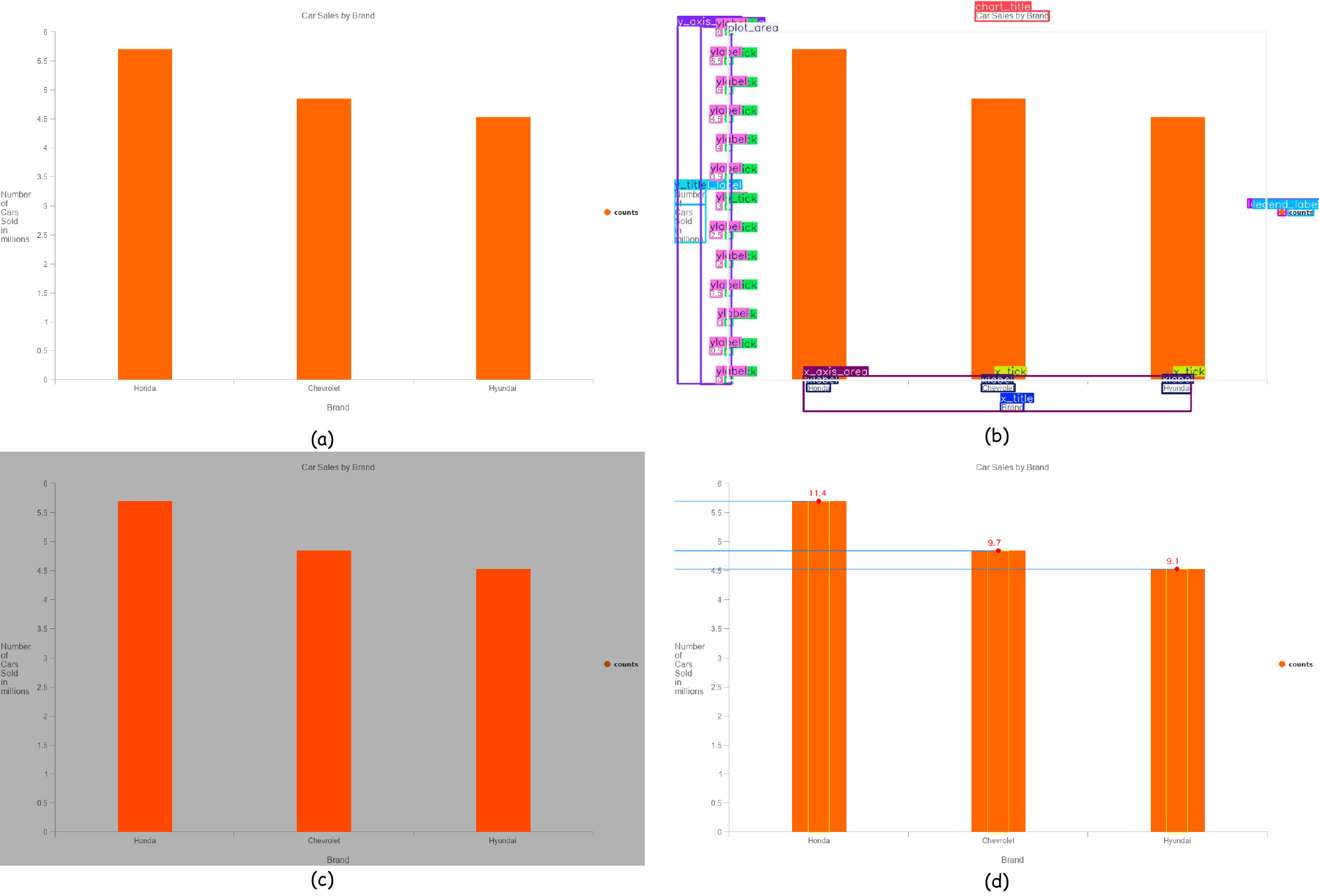}
    \caption{A typical successful case for bar chart.}
    \label{ex1}
\end{figure*}

After obtaining accurate structured visual evidence, the Reasoning Tools undertake high-level analysis and result consolidation. The workflow typically begins with a \texttt{Classification Tool} to guide subsequent orchestration. A \texttt{Relational Reasoning Tool} constructs semantic and spatial relation graphs among chart elements, while \texttt{Numerical Calculation Tool} perform ratio conversions, summations, and other arithmetic operations to mitigate the inherent computational limitations of language models. Finally, a \texttt{Data Structuring Tool} aggregates OCR text, detection boxes, auxiliary-line readings, and clustering outputs into a standardized table or JSON-based Evidence Package, achieving a tightly coupled integration of low-level perception and high-level reasoning. This pipeline enables transparent and verifiable automated analysis of complex charts. 

\section{Analysis of Cases}

\begin{figure*}[h]
    \centering
    \includegraphics[width=0.8\textwidth]{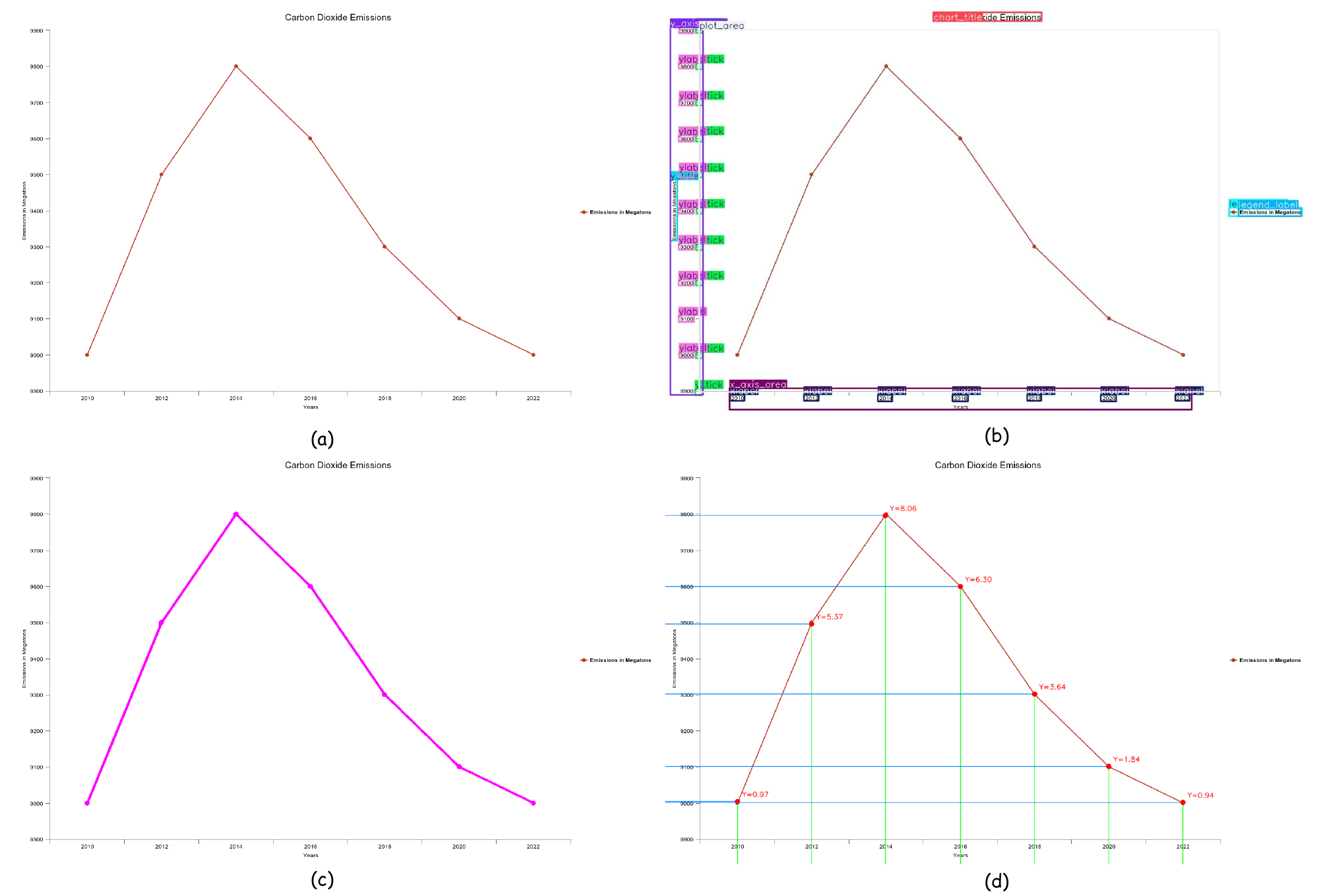}
    \caption{A typical successful case for line chart.}
    \label{ex2}
\end{figure*}

\begin{figure*}[h]
    \centering
    \includegraphics[width=\textwidth]{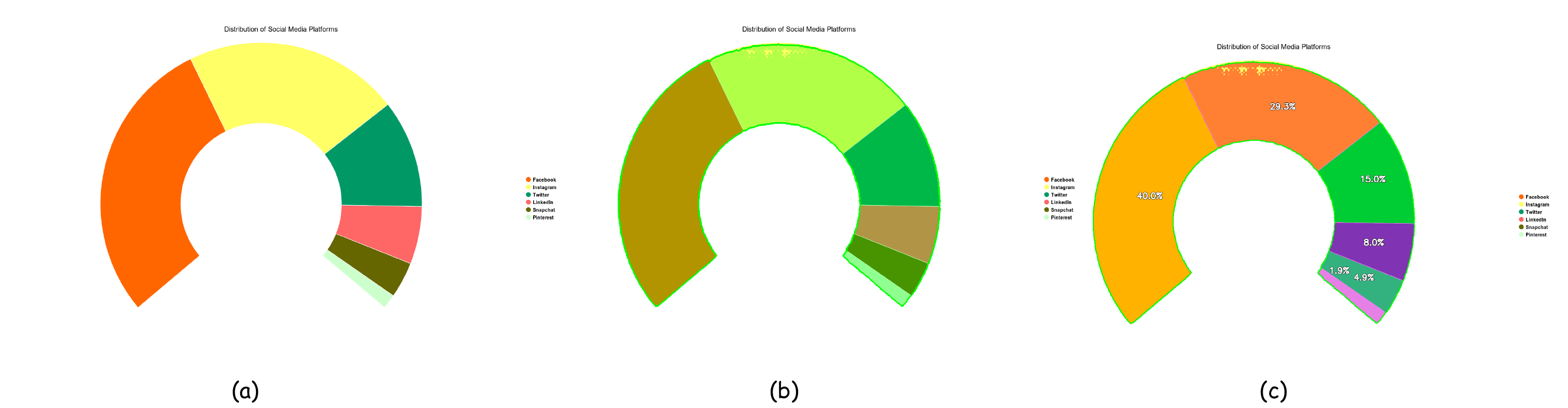}
    \caption{A typical successful case for sector charts.}
    \label{ex3}
\end{figure*}

\begin{figure*}[t]
    \centering
    \includegraphics[width=0.8\textwidth]{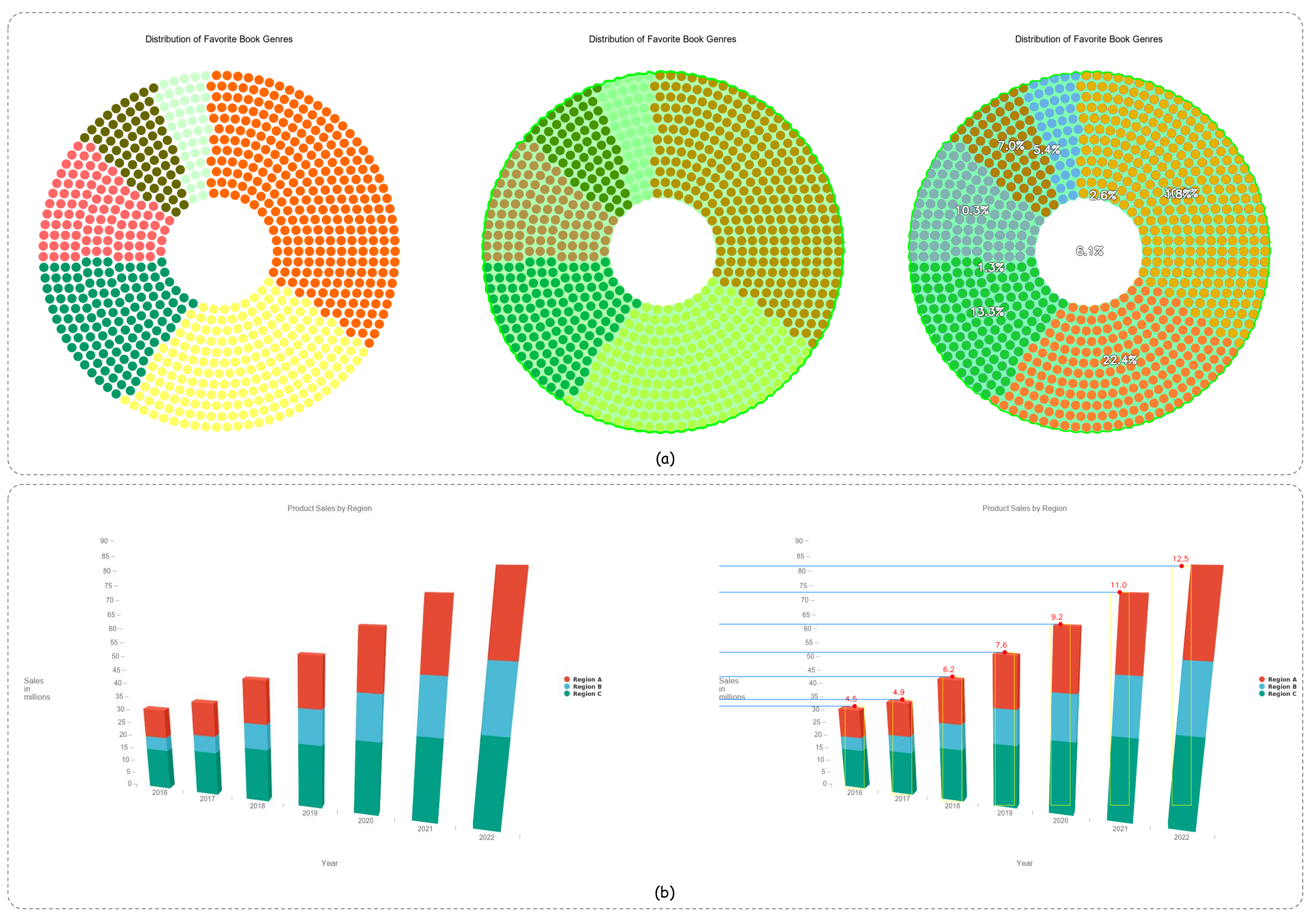}
    \caption{Some failure cases.}
    \label{ex5}
\end{figure*}

Fig.~\ref{ex1}, fig.~\ref{ex2}, and fig.~\ref{ex3} illustrate the intermediate visual outputs of ChartAgent on three common chart types—bar charts, line charts, and pie/sector charts—using samples drawn from the ChartBench test set. Fig.~\ref{ex1} presents the complete analysis pipeline for a bar chart: (a) shows the original image; (b) displays the intermediate results after key-element detection; and (c) illustrates the extraction of individual bars using the segmentation tool. Based on the information from (b) and (c), the system then draws auxiliary lines and performs numerical reading. The textual annotations beside the auxiliary lines denote relative tick counts (measured in units of the $y$-axis ticks). For the query \emph{“According to this chart, what is the Number of Cars Sold of Hyundai?”}, the ground-truth value is $4.5$. ChartAgent first estimates that the bar corresponding to Hyundai spans approximately $9.1$ tick intervals, and since each interval represents $0.5$ units, the system computes $9.1 \times 0.5 \approx 4.55$, which aligns closely with the true value.

Fig.~\ref{ex2} demonstrates the reasoning process and key intermediate artifacts for a line chart following the same workflow. Fig.~\ref{ex3} presents a sector-chart example: by sequentially invoking the segmentation tool and the enhanced AugmentSegmentation module, ChartAgent robustly performs sector instantiation and proportion estimation, effectively addressing visual-measurement scenarios that conventional MLLMs struggle to handle.

Fig.~\ref{ex5} presents several representative failure cases. In (a), the donut chart is incorrectly analyzed: unlike conventional charts with solid color fills, this example is composed of dense dotted patterns. While the primary segmentation stage remains largely correct, the enhanced segmentation procedure (segment-cluster-quantify) is adversely affected by boundary noise and color mixing introduced by the dot-matrix texture. This results in inaccurate sector instantiation and area estimation, and in some cases triggers over-segmentation or under-segmentation.

(b) shows an uncommon 3D bar chart. The auxiliary-line tool assumes a two-dimensional orthographic projection and does not model perspective or occlusion effects. Consequently, the system directly maps the visually observed 3D bar heights to numeric values, leading to systematic underestimation or overestimation.

\section{Detailed ChartAgent Algorithm}

Algorithm~\ref{alg:chartagent} details the inference procedure of \emph{ChartAgent} under a Think–Observe–Execute paradigm with an explicit computation budget. Lines 1–3 initialize the step index, the cumulative cost, the interaction history $\mathcal{H}_0$, and the evidence package $\mathcal{E}_0$. The main loop (lines 6–20) is executed while the accumulated cost $C$ remains below the budget $B$. At each iteration, ChartAgent selects an action $a_t$ from the tool library $\mathcal{T}$ or a special \texttt{<finish>} action (line 8) by maximizing a cost–benefit objective, where the expected information gain $\mathrm{EIG}(a \mid \mathcal{H}_t)$ is penalized by the tool cost $c(a)$ weighted by $\lambda$. The loop is terminated if the \texttt{finish} action is chosen or if the expected information gain of the best action falls below a threshold $\eta$ (lines 10–12), thereby preventing uninformative or redundant tool calls. Whenever a tool is executed (lines 14–19), its output $z_t$ is appended to the evidence package $\mathcal{E}_{t+1}$ and the history state $\mathcal{H}_{t+1}$ is updated via \texttt{UpdateState}, while the cumulative cost and step index are incremented accordingly.

Upon termination of the Think–Observe–Execute loop, ChartAgent proceeds to a reflection phase driven by multi-expert collaboration (lines 22–24). In this stage, the procedure \texttt{GroupTalk} takes as input the chart image $x$, the user query $q$, the final history $\mathcal{H}_t$, and the evidence package $\mathcal{E}_t$ to obtain a set of expert votes, denoted as \textit{Votes}. These votes encode the independent assessments of multiple specialized expert modules conditioned on the same collection of intermediate reasoning steps and tool outputs. The function \texttt{AggregateAndCalibrate} then fuses and calibrates these votes to produce the final answer $y$, which is returned together with the final evidence package $\mathcal{E}_t$. This two-stage design—cost-aware iterative evidence acquisition followed by collaborative reflection—enables ChartAgent to balance computational efficiency, predictive accuracy, and interpretability in chart understanding tasks.~\cite{scicqa}

\begin{algorithm}[t!]
\small
\caption{ChartAgent Inference Framework}
\label{alg:chartagent}
\KwIn{Chart image $x$; user instruction $q$; tool library $\mathcal{T}$; budget $B$.}
\KwOut{Final answer $y$; evidence package $\mathcal{E}$.}

\BlankLine
\tcp{Initialization}
$t \gets 0$; $C \gets 0$ \tcp*[r]{Step index and current cost}
$\mathcal{H}_0 \gets \emptyset$; $\mathcal{E}_0 \gets \emptyset$ \tcp*[r]{History and evidence}

\BlankLine
\tcp{Think-Observe-Execute loop}
\While{$C < B$}{
    \tcp{Select action via cost--gain trade-off}
    $a_t \gets \arg\max\limits_{a \in \mathcal{T} \cup \{\text{finish}\}}
        \big( \mathrm{EIG}(a \mid \mathcal{H}_t) - \lambda \cdot c(a) \big)$\;

    \If{$a_t = \text{finish}$ \textbf{or} $\mathrm{EIG}(a_t) \le \eta$}{
        \textbf{break}\;
    }

    \tcp{Execute tool and update evidence}
    $z_t \gets \Execute(a_t, x, \mathcal{H}_t)$\;
    $\mathcal{E}_{t+1} \gets \mathcal{E}_t \cup \{(a_t, z_t)\}$\;
    $\mathcal{H}_{t+1} \gets \UpdateState(\mathcal{H}_t, z_t)$\;
    $C \gets C + c(a_t)$\;
    $t \gets t + 1$\;
}

\BlankLine
\tcp{Reflect: multi-expert collaboration}
$Votes \gets \GroupTalk(x, q, \mathcal{H}_t, \mathcal{E}_t)$\;
$y \gets \AggregateAndCalibrate(Votes)$\;

\Return{$y, \mathcal{E}_t$}\;
\end{algorithm}

%
%
%

\begin{figure*}[h!]
    \centering
    \includegraphics[width=\textwidth]{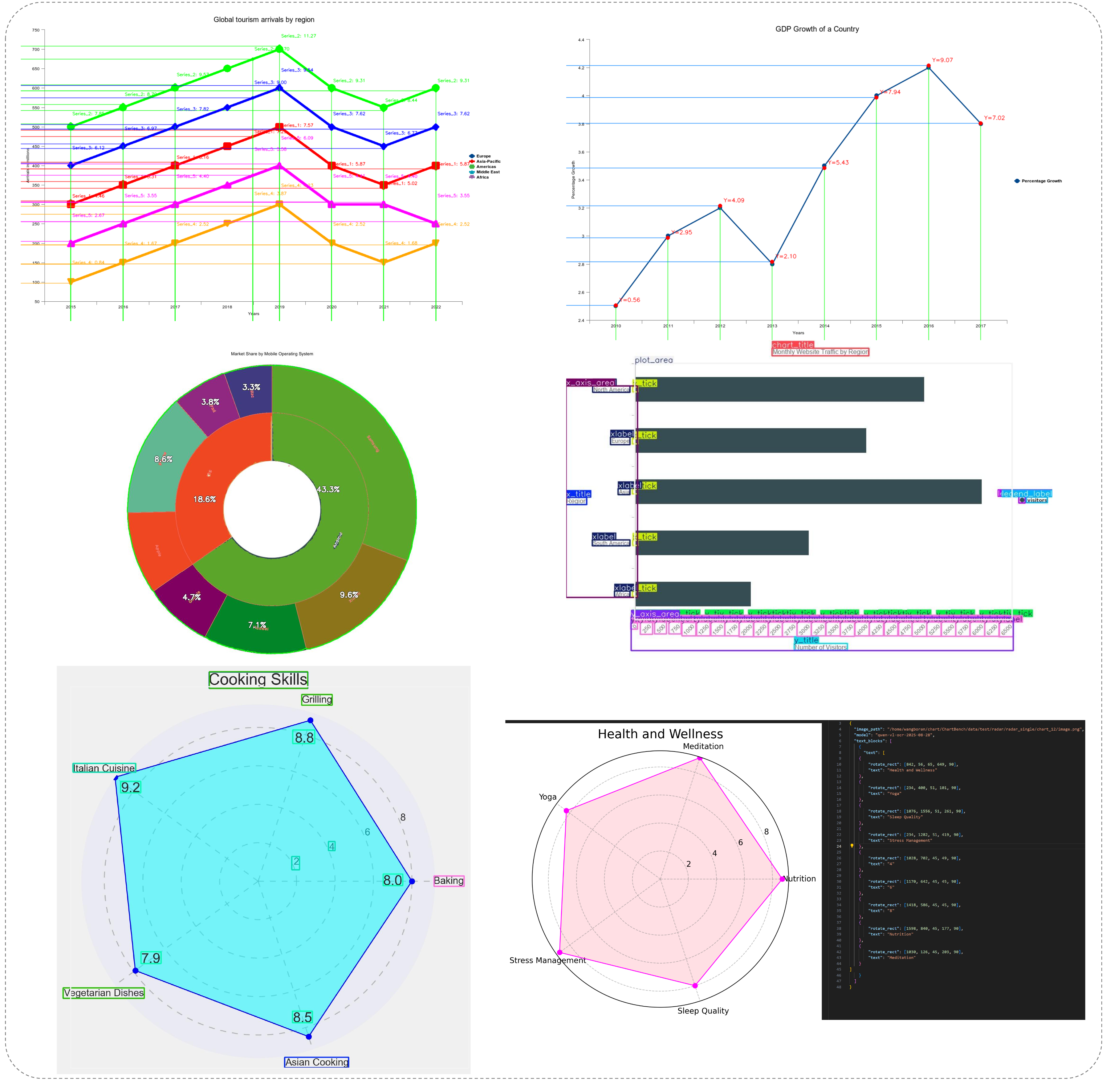}
    \caption{Several representative examples of CEP.}
    \label{ex6}
\end{figure*}


\end{document}